\pdfoutput=1

\documentclass[11pt]{article}

\usepackage[final]{coling}

\usepackage{times}
\usepackage{latexsym}

\usepackage[T1]{fontenc}

\usepackage[utf8]{inputenc}

\usepackage{microtype}

\usepackage{inconsolata}

\usepackage{graphicx}

\usepackage[utf8]{inputenc}
\usepackage{CJKutf8}
\usepackage[whole]{bxcjkjatype}
\usepackage{tikz}
\newcommand{\bgfa}[1]{\tikz[baseline=(X.base)]{\node(X)[rectangle, fill=cyan!60!blue!18, rounded corners, text height=.8ex,text depth=-0.5ex]{\textit{#1}};}}
\newcommand{\bgfb}[1]{\tikz[baseline=(X.base)]{\node(X)[rectangle, fill=red!60!blue!18, rounded corners, text height=.8ex,text depth=-0.5ex]{\textit{#1}};}}
\newcommand{\bgfc}[1]{\tikz[baseline=(X.base)]{\node(X)[rectangle, fill=green!60!blue!18, rounded corners, text height=.8ex,text depth=-0.5ex]{\textit{#1}};}}

\usepackage{heatmap}

\usepackage{tablefootnote}

\title{LCTG Bench: LLM Controlled Text Generation Benchmark}

\author{
 \textbf{Kentaro Kurihara\textsuperscript{1,2}},
 \textbf{Masato Mita\textsuperscript{2}},
 \textbf{Peinan Zhang\textsuperscript{2}},
 \textbf{Shota Sasaki\textsuperscript{2}},
\\
 \textbf{Ryosuke Ishigami\textsuperscript{2}},
 \textbf{Naoaki Okazaki\textsuperscript{3}},
\\
 \textsuperscript{1}AI Shift Inc.,
 \textsuperscript{2}CyberAgent Inc.,
 \textsuperscript{3}Tokyo Institute of Technology,
\\
 \small{
   {kurihara\_kentaro, mita\_masato, zhang\_peinan, sasaki\_shota, ishigami\_ryosuke@cyberagent.co.jp}
 }
 \\
 \small{
   {okazaki@c.titech.ac.jp}
 }
}

\begin{document}
\maketitle
\begin{abstract}
The rise of large language models (LLMs) has led to more diverse and higher-quality machine-generated text. 
However, their high expressive power makes it difficult to control outputs based on specific business instructions.
In response, benchmarks focusing on the controllability of LLMs have been developed, but several issues remain: (1) They primarily cover major languages like English and Chinese, neglecting low-resource languages like Japanese; (2) Current benchmarks employ task-specific evaluation metrics, lacking a unified framework for selecting models based on controllability across different use cases.
To address these challenges, this research introduces LCTG Bench, the first Japanese benchmark for evaluating the controllability of LLMs.
LCTG Bench provides a unified framework for assessing control performance, enabling users to select the most suitable model for their use cases based on controllability. 
By evaluating nine diverse Japanese-specific and multilingual LLMs like GPT-4~\citep{OpenAI_GPT4_2023}, we highlight the current state and challenges of controllability in Japanese LLMs and reveal the significant gap between multilingual models and Japanese-specific models.

\end{abstract}

\section{Introduction}
\label{sec:introduction}

\begin{figure*}[ht!]
\begin{center}
\includegraphics[width=0.99\linewidth]{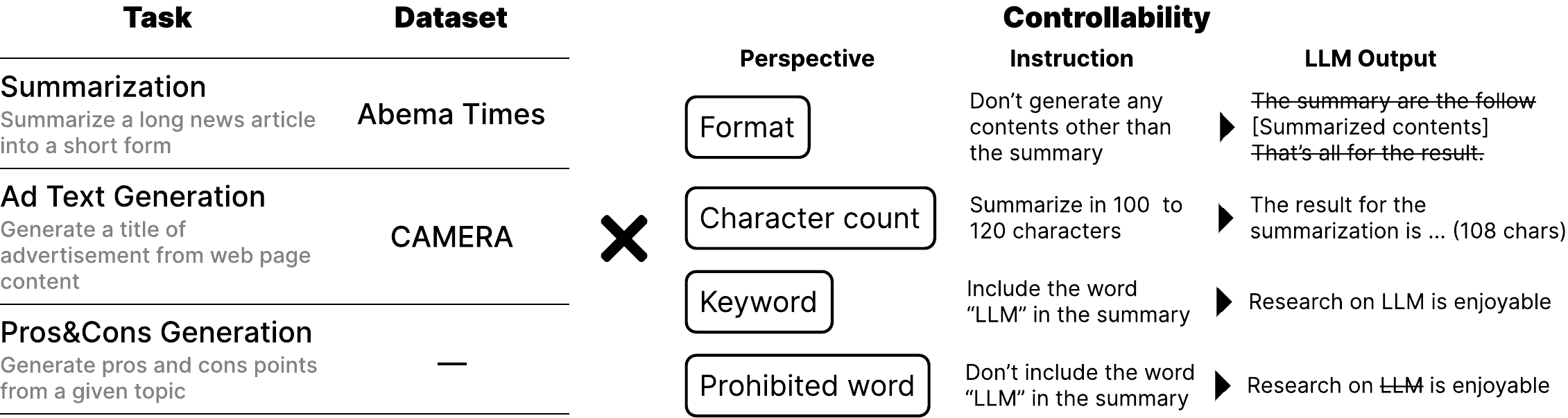}
\end{center}
\caption{Overview of LCTG Bench.}
\label{fig:ponchi_figure}
\end{figure*}

Research and development of large language models (LLMs) have accelerated worldwide since the release of OpenAI's ChatGPT.
Evaluating the abilities of the LLM from multiple viewpoints is essential to develop a high-performing LLM and apply it to business scenarios.
The performance of LLMs can be evaluated from different perspectives using a variety of benchmarks, including questions related to knowledge~\citep{hendrycks2021measuring} and questions that are difficult for the LLM~\citep{suzgun-etal-2023-challenging}.
 
However, applying LLMs to real-world business scenarios requires evaluating not only their knowledge, problem-solving abilities, fluency, and consistency but also their controllability---specifically, how well they generate text that adheres to given instructions.
For example, it is necessary to follow submission rules or SEO strategies (e.g., specific character counts, specific word constraints, and formatting) when using an LLM to create news articles and online advertisements.
In response to this, benchmarks focusing on the controllability of LLMs have been developed, but there are still some issues: (1) These mainly target major languages such as English~\citep{zhou2023instructionfollowing} and Chinese~\citep{he2024celloAAAI, jing2023followeval}, and do not cover low-resource languages such as Japanese; (2) The current benchmarks use task-specific evaluation metrics~\citep{sun2023evaluating} , and there is a lack of a unified framework for selecting models based on controllability in different use cases.

To address these limitations, this study introduces \textsc{LCTG Bench} (\textbf{L}LM \textbf{C}ontrolled \textbf{T}ext \textbf{G}eneration \textbf{Bench}mark)
a benchmark for evaluating the controllability of LLMs, assuming their practical application (Figure \ref{fig:ponchi_figure}). 
The LCTG Bench comprises three generative tasks: \textit{Summarization}, \textit{Ad Text Generation}, and \textit{Pros \& Cons Generation}, each with distinct characteristics, allowing for the evaluation of the controllability performance of LLMs from four verifiable rule-based perspectives: \textit{Format}, \textit{Character count}, \textit{Keyword}, \textit{Prohibited word}, which we select from considering the adaption of LLM in the real-world.

There are two contributions of the LCTG Bench. 
Firstly, it is the first benchmark in Japanese that focuses on the controllability of LLMs, and we evaluate various Japanese LLMs, including multilingual models like GPT-3.5, GPT-4, and Gemini~\citep{geminiteam2024geminifamilyhighlycapable}.
Secondly, it enables to evaluate the controllability of LLMs under the same constraints across all tasks.
This evaluation framework provides robustness that allows the selection of the best model for each use case in terms of controllability.
For evaluating controllability in this study, we also evaluated the generated text using GPT-4 to confirm that the generated text meets the task requirements to some extent.
The evaluation of various LLMs using LCTG Bench shows the current status and issues in the controlled text generation of Japanese LLMs, including the large gap between GPT-4 and Japanese LLMs.

\section{Related Work}
\label{sec:related_work}
In English and Chinese, there has been a movement to examine the controllability of LLMs. 
For English, \citet{sun2023evaluating} constructed a benchmark \textsc{NPB} focusing on numerical constraints, and \citet{zhou2023instructionfollowing} introduced a benchmark \textsc{IFEval} and evaluated controllability from various perspectives such as keyword and format. 
~\citet{yao2023collie} have conducted a rule-based LLM evaluation of instructional adaptability regarding specific letters, number of words, etc.
\citet{liu2023benchmarking} analyzed the quality of summary generation results and conducted evaluation by using LLMs for scoring~ method\citep{fu2023gptscore} and ranking method~\citep{sun2023chatgpt}.
Also, \citet{jiang2023followbench} have conducted rule-based and LLM-based evaluations.
Furthermore, not only in English but also in Chinese, the English-Chinese benchmark \textsc{FollowEval}, which emphasizes manual construction and rule-based evaluation, has been released.
However, these studies \textbf{either evaluate controllability differently for each task or use mixed data across multiple tasks}, failing to offer a robust framework for selecting the most suitable model for each use case in terms of controllability.
Recently, \citet{he2024celloAAAI} released \textsc{CELLO}, a benchmark for evaluating cross-task controllability.
Although this work shares the same objective as ours, there are some differences, such as the items to be controlled (with or without negation-related items) and the method of guaranteeing the quality of generated output (surface information-based or semantic information-based method). 
CELLO does not adopt a negative evaluation perspective and uses surface information-based quality assurance. 
Most importantly, since CELLO is a Chinese dataset, our LCTG Bench (Japanese dataset) complements their dataset (and vice versa).

As mentioned above, there is no evaluation benchmarks on the controllability of LLMs in Japanese, and existing research evaluates LLMs by developing Japanese versions of natural language understanding (NLU) benchmarks and datasets~\cite{kurihara-etal-2022-jglue, tikhonov-ryabinin-2021-heads, someya2023jcola, Kurihara_nlp2020,shi2022language,hasan-etal-2021-xl} such as NLI and QA, and integrated leaderboards for LLMs such as \texttt{lm-evaluation-harness}~\cite{eval-harness}\footnote{\url{https://github.com/Stability-AI/lm-evaluation-harness/tree/jp-stable}} and \texttt{MT-Bench}~\cite{zheng2023judging}\footnote{\url{https://github.com/Stability-AI/FastChat/tree/jp-stable/fastchat/llm_judge}}.

\section{LCTG Bench}
\label{ssec:benchmark}

\begin{table*}[!h]
\small
\begin{center}
\begin{tabular}{ccccccc}
\hline
\multicolumn{1}{c}{Task} & Dataset & \multicolumn{1}{c}{\textsc{Format}} & \multicolumn{1}{c}{\textsc{C-count}} & \multicolumn{1}{c}{\textsc{Keyword}} & \multicolumn{1}{c}{\textsc{P-word}} & \multicolumn{1}{c}{Avg.char count} \\ \hline
Summarization & ABEMA TIMES & 120 & 120 & 120 & 120 & 726\\
Ad Text Generation & CAMERA & 150 & 150 & 150 & 150 & 184\\
Pros \& Cons Generation & - & 150 & 150 & 150 & 150 & 85\\ 
\hline
\end{tabular}
\caption{LCTG Bench statistics. Avg.char count means the average number of characters in the sample for each task.}
\label{tab:benchmark_statistics}
\end{center}
\end{table*}




\begin{figure*}[t]
 \centering
  \includegraphics[width=0.9\linewidth]{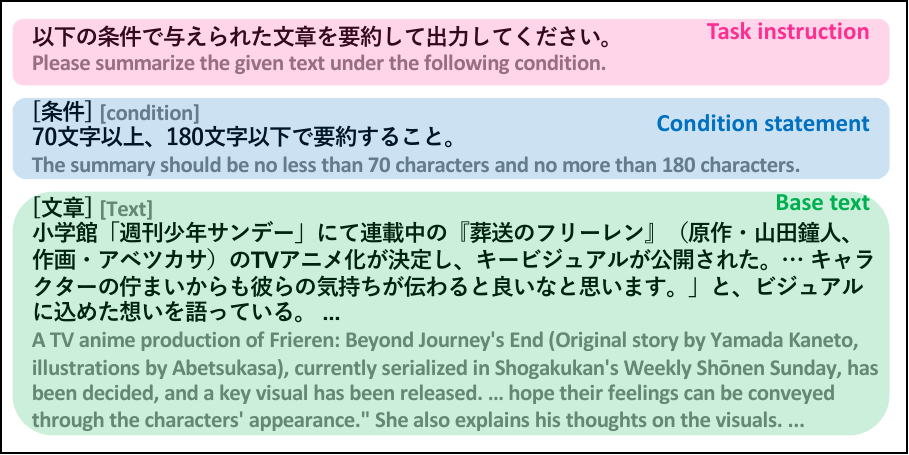}
 \caption{An example of summarization prompt (Character count).}
 \vspace{-3mm}
 \label{fig:summary_sample}
\end{figure*}



\begin{figure*}[t]
 \centering
  \includegraphics[width=0.9\linewidth]{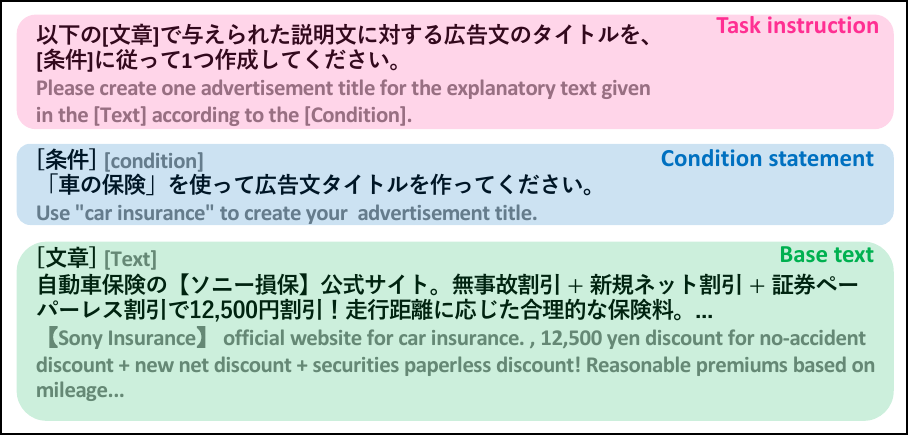}
 \caption{An example of ad text generation prompt (Keyword).}
 \label{fig:ad_generation_sample}
\end{figure*}


\begin{figure*}[t]
 \centering
  \includegraphics[width=0.9\linewidth]{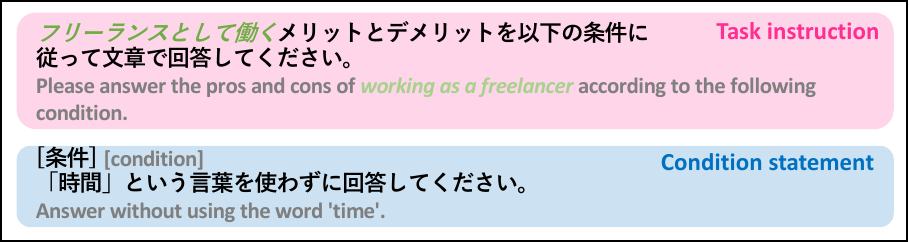}
 \caption{An example of pros \& cons generation prompt (Prohibited word). In this example, ``フリーランスとして働く (working as a freelancer)'' is a pros \& cons topic.}
 \label{fig:merideme_generation_sample}
\end{figure*}

This study aims to create a Japanese benchmark that offers guidelines to help developers select the appropriate model for each use case in real-world scenarios, focusing specifically on controllability (\S\ref{sec:introduction}).
In constructing such a benchmark, we first defined the following two design requirements: (1) The tasks must align with real-world scenarios where controllability is important, and (2) each control item must be able to be evaluated across tasks.
Requirement 2 assumes that given the wide variety of real-world use cases for LLMs, a framework that evaluates controllability across different tasks can offer transferable guidelines applicable to a broad range of scenarios.

To meet Requirement 1, we select the following three generative tasks for this study: \textit{Summarization}, \textit{Ad Text Generation}, \textit{Pros \& Cons Generation}. 
In addition, we select the following four evaluation perspectives of controllability to satisfy Requirement 2: Format (\textsc{Format}), Character count (\textsc{C-count}), Keyword (\textsc{Keyword}), and Prohibited word (\textsc{P-word}), that need to be evaluated in consideration of actual business scenarios, such as submitting articles and online advertisements generated by LLM or incorporating LLM into a system.

Also, to compare the difficulty of controllability between tasks, this benchmark evaluates them from unified four controllability perspectives, regardless of the task.
The details of each task and the perspectives of controllability are described in \S\ref{ssec:benchmark:task_overview} and \S\ref{ssec:benchmark:eval_item}, respectively.
We show the statistics of the LCTG Bench in Table \ref{tab:benchmark_statistics}.

\subsection{Overview of how each sample is constructed}

Considering the substantial generative capabilities of LLMs, we posit that the current approach of evaluating generated texts, which is based on a predetermined correct generative result, may not provide an essential evaluation~\citep{maynez-etal-2023-benchmarking}. 
Therefore, we prepare the input without a gold reference in this benchmark, providing a rule-based verification of LLMs' controllability performance without compromising their diversity of generation capabilities. 
The examples of the prompts for the three tasks are shown in Figures \ref{fig:summary_sample}, \ref{fig:ad_generation_sample}, and \ref{fig:merideme_generation_sample}. 
Each prompt comprises three elements: `\bgfb{task instruction}', `\bgfa{condition statement}', and `\bgfc{the base text of the task}'\footnote{The Pros \& Cons Generation has no base text; instead, the pros \& cons topic is in the task instruction.}. 
We can evaluate various controllability performances using the same task instructions and base texts by changing the condition statements.

Also, even with the same meaning, the model's output may vary if the prompt has different expressions~\citep{mizrahi2023state}. 
Therefore, to include condition statements with diverse expressions within the same perspectives of controllability, we collect templates for condition statements other than `Format' using crowdsourcing
\footnote{We used Yahoo! Crowdsourcing (\url{https://crowdsourcing.yahoo.co.jp/}).}. 
The values to be substituted into the templates, such as numbers and words, are collected differently for each task.

The base text for the task target is collected from publicly available datasets and data our organization can disclose. 
The method and examples of collecting templates by using crowdsourcing are shown in the Appendix \ref{appendix:cs_example}.

\subsection{Task Overview and Procedure of Condition Statements}
\label{ssec:benchmark:task_overview}
\paragraph{Summarization}
We adopt summarization, which is a complex and widely applied language-generation task in business. 
In summarization, as depicted in Figure \ref{fig:summary_sample}, we construct a dataset of prompt sets that instruct the summarization of sentences based on the condition statement. We used 120 articles from the news site ``ABEMA TIMES''
\footnote{\url{https://times.abema.tv/}}
\footnote{Due to public constraints, we selected six categories from categories other than `News': ``Entertainment'', ``Sports'', ``Anime'', ``Shogi'', ``Mahjong'', and ``HIPHOP''} as the base text for summarization. 
Considering the values for the condition statement templates, we set the value to the \textsc{C-count} templates, with the maximum and minimum limits of 200 characters and 50 characters, respectively.
We assign the values randomly in multiples of 10. 
Considering \textsc{Keyword}, keywords are extracted from the summary results generated by GPT-4 as the crucial word in articles. 
Specifically, we generate five summaries for each sample article using GPT-4. 
Then, the words appearing commonly in the resulting summary set are extracted as high-importance words and designated as keywords. 
This method is based on the assumption that words frequently appearing in summaries generated by high-performance LLMs like GPT-4 are highly important in the document. 
As for \textsc{P-word} while it may be rare for high-importance words to be designated as prohibited words in actual summarization scenarios, we select them from the high-importance words in the context of this task to measure the ability to output expressions excluding the specified ones, in an identical method to how keywords are determined.

\paragraph{Ad Text Generation}
We introduce the ad text generation task, which requires fewer output characters than the summarization and is highly important for \textsc{Keyword} and \textsc{P-word}. In the ad text generation, as shown in Figure \ref{fig:ad_generation_sample}, we construct a dataset of prompt sets instructing to create a title of ad text according to the condition statement. 
The base texts for generating ad text titles is collected from the descriptions of landing pages (LP) of cases with two search keywords from the evaluation data of the ad text generation benchmark CAMERA~\citep{mita2023}\footnote{\url{https://github.com/CyberAgentAILab/camera}}.
For the \textsc{C-count}, while setting upper and lower limits as in the summarization, considering that it is a task that expects output in a smaller number of characters compared to the summarization, we substitute random multiples of 5 between the upper limit of 50 characters and the lower limit of 20 characters. 
For \textsc{Keyword}/ \textsc{P-word}, we use the search keywords given to the samples in CAMERA.

\paragraph{Pros \& Cons Generation}
We introduce a task that involves freely writing about the pros \& cons of a given topic. This task differs from summarization and ad text generation, requiring generating text based on a topic in the instruction. Understanding the pros \& cons is essential in enabling the LLM to conduct discussions and evaluations on behalf of humans~\citep{chan2023chateval}. 
Therefore, this task was adopted. 
To collect candidate topics for pros \& cons generation, we used the GPT-4 and collected 150 cases manually.
Like in the summarization, we select two words of high importance in discussing pros \& cons as \textsc{Keyword} and `\textsc{P-word}'. 
As for the \textsc{C-count}, a value that is a multiple of 10 is randomly assigned. 
The total length of the combined pros \& cons is a minimum of 100 characters and a maximum of 400 characters.

\subsection{Perspectives on Controllability of LLM}
\label{ssec:benchmark:eval_item}

The LCTG Bench evaluates ``whether the generated text satisfies the conditions'' in each of the four constraints, respectively.
\paragraph{Format (\textsc{Format})}
In actual business situations, such as integrating LLM into a system, the control performance of LLM's Format is required, but it has been pointed out that LLM often fails to comply with instructions regarding the format of its output.
Therefore, tools such as `Function Calling'\footnote{\url{https://platform.openai.com/docs/guides/function-calling}} are sometimes used for format shaping, but their accuracy is imperfect.
In this study, we evaluate the essential performance in terms of format as ``the ability to generate without adding unnecessary explanatory sentences before and after the output.'' 
We create condition statements (Appendix \ref{appendix:format_prompt_ex}) instructing not to add sentences other than the required content.

\paragraph{Character Count (\textsc{C-count})}
The number of characters in business applications such as article and the title of advertisement creation may be restricted. 
This study evaluates whether the LLM can generate text according to a specified character count within a given range.

\paragraph{Keyword (\textsc{Keyword}) / Prohibited Word (\textsc{P-word})}
In actual use cases, such as creating a website title or tagline or applying LLM to a chatbot, it may be desirable to specify keywords or prohibited words considering public order and morals, SEO measures, and so on.
This study evaluates the system's ability to generate sentences incorporating keywords and prohibited words. 

\section{Evaluation of LLM using LCTG Bench}

We conducted an evaluation experiment of the Japanese LLMs using the LCTG Bench to demonstrate their current status and issues of them and the utility of this benchmark. 

The models used in the experiment were selected to encompass a range of types and parameter quantities, including high-performance models such as GPT-4 and base models like Llama 2~\citep{touvron2023llama} and GPT-NeoX~\citep{black2022gptneox20b}. 
The hyperparameters and system prompts for the various LLMs were used at their default values based on the values listed on the Hugging Face Hub.
The models used in the experiment and the settings of hyperparameters are shown in the Appendix \ref{appendix:hypara}.

\subsection{Evaluation Settings}
Focusing only on controllability may cause us to overlook samples whose content deviates significantly from the task requirements while the condition is met.
Therefore, it is necessary to evaluate the quality of the generated results in addition to controllability. 
Moreover, LLMs may produce different outputs even when given the same prompt~\citep{ouyang2023llm}, this leads to variability in its evaluation results. 
Thus, we generate the output three times for each prompt and consider the average score of each time as the final score.

In addition, the text generated by the LLM may contain explanatory text at the beginning or end of sentences unrelated to the current task, and the presence or absence of irrelevant explanatory text is judged in the \textsc{Format} constraints.
However, in the context of the other three aspects of controllability and the quality evaluation of the generated text, the inclusion of unrelated explanatory text could potentially obstruct a fair evaluation of the task-specific output (refer to the Appendix \ref{appendix:with_hf_example}).
Therefore, the controllability evaluation from the three perspectives other than \textsc{Format} and the quality evaluation of the generated results in all four perspectives are performed on the generated results after removing the unnecessary explanatory text from the LLM output using GPT-4\footnote{The prompt used for the removal of unnecessary explanatory text is shown in the Appendix \ref{appendix:hf_remover_prompt}.}.
The controllability of \textsc{Format} is evaluated by comparing the generated results before and after the removal work by GPT-4.

\paragraph{Controllability Evaluation}
For the \textsc{Format}, we compare the results generated before and after unnecessary explanatory texts are removed by GPT-4.
In the summarization and pros \& cons generation, we count the cases where the beginning and the last ten characters of two sentences exactly match, to get the score.
Also, in the ad text generation task, we calculate the percentage of cases where the beginning and the last five characters of the two sentences exactly match\footnote{The removal operation by GPT-4 may change the middle part of the generated result. Therefore, it is difficult to determine whether or not there are unnecessary explanatory sentences based on an exact match between the generated results before and after the removal operation. So, we compare the strings at the beginning and end of the two sentences.}.
The number of characters to be compared is determined based on the number of characters handled in each task.
For the \textsc{C-count}, we calculate the samples where the generated text falls within the character count range specified in the condition. 
For \textsc{Keyword} and \textsc{P-word}, we determine the instances where the generated text includes (or excludes) the words specified in the condition.

\paragraph{Quality Evaluation}
For the quality evaluation of the generated text, we use GPT-4 as the evaluator. 
Specifically, we utilized GPT-4 to conduct a classification task for the quality assurance of the generated text by using the prompts shown in the Appendix \ref{appendix:quality_check_prompt}. 
To get the score, we count the cases of the appropriate generated text.

\subsection{Results}
\label{sec:results}

\begin{table*}[!t]
\fontsize{10.6pt}{10.6pt}\selectfont
\begin{center}
\begin{tabular*}{\textwidth}{@{}lV@{}VV@{}VV@{}VV@{}V|V@{}V}
\hline
\multicolumn{1}{c}{}&\multicolumn{2}{c}{\begin{tabular}[c]{@{}c@{}}\textsc{Format}\end{tabular}}&\multicolumn{2}{c}{\begin{tabular}[c]{@{}c@{}}\textsc{C-count}\end{tabular}}&\multicolumn{2}{c}{\begin{tabular}[c]{@{}c@{}} \textsc{Keyword}\end{tabular}}&\multicolumn{2}{c|}{\begin{tabular}[c]{@{}c@{}} \textsc{P-word}\end{tabular}}&\multicolumn{2}{c}{\begin{tabular}[c]{@{}c@{}}Average\end{tabular}} \\
\multicolumn{1}{c}{Model}&\multicolumn{1}{c}{CTG}&\multicolumn{1}{c}{Qual}&\multicolumn{1}{c}{ CTG}&\multicolumn{1}{c}{Qual}&\multicolumn{1}{c}{CTG}&\multicolumn{1}{c}{Qual}&\multicolumn{1}{c}{CTG}&\multicolumn{1}{c|}{Qual}&\multicolumn{1}{c}{CTG}&\multicolumn{1}{c}{Qual}\\ \hline
gpt-4-1106-preview & \multicolumn{1}{W@{}}{0.992}   & \multicolumn{1}{@{}W}{0.925} & 0.450 & 0.869  & \multicolumn{1}{W@{}}{0.972} & \multicolumn{1}{@{}W}{0.886}  & \multicolumn{1}{W@{}}{0.970} & \multicolumn{1}{@{}W|}{0.775} & \multicolumn{1}{W@{}}{0.846} & \multicolumn{1}{@{}W}{0.864}   \\
gpt-3.5-turbo-0125 & 0.978 & 0.883  & 0.478 & 0.853  & 0.970 & 0.811 & 0.450 & 0.828   & 0.719 & 0.844  \\
gemini-pro  & 0.914   & 0.894 & \multicolumn{1}{W@{}}{0.486} & \multicolumn{1}{@{}W}{0.881} & 0.939  & 0.856     & 0.645  & 0.817  & 0.746  & 0.862 \\\hline
ca/calm2-7b-chat & 0.881   & 0.486   & 0.219 & 0.428   & 0.808  & 0.444  & 0.303  & 0.403   & 0.553   & 0.440   \\
elyza/llama2-7b-instr & 0.458   & 0.789   & 0.325 & 0.792 & 0.803  & 0.806  & 0.305  & 0.778  & 0.473   & 0.791 \\
line/llm-3.6b-instr      & 0.344   & 0.067 & 0.125 & 0.055 & 0.597  & 0.044  & 0.525  & 0.056   & 0.398   & 0.056 \\
matsuo/wl-10b-instr & 0.944   & 0.530 & 0.194 & 0.500 & 0.614  & 0.458 & 0.500  & 0.495 & 0.563   & 0.496    \\
rinna/youri-7b-chat & 0.911   & 0.675  & 0.166 & 0.683 & 0.647  & 0.609 & 0.492  & 0.611 & 0.554  & 0.645  \\
stabilityai/lm-instr-7b      & 0.847   & 0.603 & 0.200 & 0.517 & 0.706  & 0.517 & 0.386  & 0.532 & 0.535   & 0.542 \\
\hline
\end{tabular*}
\caption{Evaluation results of controllability (CTG) and quality (Qual) of generation in \textbf{summarization}.}
\label{tab:summary-quality-result}
\end{center}
\end{table*}

\begin{table*}[!t]
\small
\fontsize{10.6pt}{10.6pt}\selectfont
\begin{center}
\begin{tabular*}{\textwidth}{@{}lV@{}VV@{}VV@{}VV@{}V|V@{}V}
\hline
\multicolumn{1}{c}{}&\multicolumn{2}{c}{\begin{tabular}[c]{@{}c@{}}\textsc{Format}\end{tabular}}&\multicolumn{2}{c}{\begin{tabular}[c]{@{}c@{}}\textsc{C-count}\end{tabular}}&\multicolumn{2}{c}{\begin{tabular}[c]{@{}c@{}} \textsc{Keyword}\end{tabular}}&\multicolumn{2}{c|}{\begin{tabular}[c]{@{}c@{}} \textsc{P-word}\end{tabular}}&\multicolumn{2}{c}{\begin{tabular}[c]{@{}c@{}}Average\end{tabular}} \\
\multicolumn{1}{c}{Model}&\multicolumn{1}{c}{CTG}&\multicolumn{1}{c}{Qual}&\multicolumn{1}{c}{ CTG}&\multicolumn{1}{c}{Qual}&\multicolumn{1}{c}{CTG}&\multicolumn{1}{c}{Qual}&\multicolumn{1}{c}{CTG}&\multicolumn{1}{c|}{Qual}&\multicolumn{1}{c}{CTG}&\multicolumn{1}{c}{Qual}\\ \hline
gpt-4-1106-preview & 0.971   & \multicolumn{1}{@{}W}{0.984}     & 0.358 & \multicolumn{1}{@{}W}{0.982}   & \multicolumn{1}{W@{}}{0.813} & \multicolumn{1}{@{}W}{0.736} & \multicolumn{1}{W@{}}{1.000} & \multicolumn{1}{@{}W|}{0.951}  & \multicolumn{1}{W@{}}{0.786}   & \multicolumn{1}{@{}W}{0.913} \\
gpt-3.5-turbo-0125 & 0.842 & 0.847   & \multicolumn{1}{W@{}}{0.480} & 0.944    & 0.787 & 0.640  & 0.973 & 0.742    & 0.771  & 0.793    \\
gemini-pro  & \multicolumn{1}{W@{}}{0.976}   & 0.824 & 0.431 & 0.889  & 0.718  & 0.584   & 0.925  & 0.756 & 0.762  & 0.763 \\\hline
ca/calm2-7b-chat      & 0.864   & 0.753  & 0.414 & 0.800 & 0.389  & 0.680  & 0.678  & 0.634  & 0.586   & 0.717 \\
elyza/llama2-7b-instr  & 0.813   & 0.562   & 0.260 & 0.804 & 0.671  & 0.500  & 0.791  & 0.711 & 0.634   & 0.644 \\
line/llm-3.6b-instr      & 0.722   & 0.395  & 0.131 & 0.351  & 0.364  & 0.282  & 0.704  & 0.280 & 0.481   & 0.327  \\
matsuo/wl-10b-instr & 0.836   & 0.642  & 0.242 & 0.653  & 0.336  & 0.551  & 0.760  & 0.631  & 0.543   & 0.620 \\
rinna/youri-7b-chat & 0.800   & 0.320  & 0.182 & 0.365 & 0.616  & 0.265 & 0.662  & 0.278 & 0.565   & 0.307  \\
stabilityai/lm-instr-7b      & 0.529   & 0.504 & 0.271 & 0.551 & 0.593  & 0.380 & 0.698  & 0.498 & 0.523 & 0.483  \\
\hline
\end{tabular*}
\caption{Evaluation results of controllability (CTG) and quality (Qual) of generation in \textbf{ad text generation}.}
\label{tab:ad-generation-quality-result}
\end{center}
\end{table*}

\begin{table*}[!t]
\fontsize{10.6pt}{10.6pt}\selectfont
\begin{center}
\begin{tabular*}{\textwidth}{@{}lV@{}VV@{}VV@{}VV@{}V|V@{}V}
\hline
\multicolumn{1}{c}{}&\multicolumn{2}{c}{\begin{tabular}[c]{@{}c@{}}\textsc{Format}\end{tabular}}&\multicolumn{2}{c}{\begin{tabular}[c]{@{}c@{}}\textsc{C-count}\end{tabular}}&\multicolumn{2}{c}{\begin{tabular}[c]{@{}c@{}} \textsc{Keyword}\end{tabular}}&\multicolumn{2}{c|}{\begin{tabular}[c]{@{}c@{}} \textsc{P-word}\end{tabular}}&\multicolumn{2}{c}{\begin{tabular}[c]{@{}c@{}}Average\end{tabular}} \\
\multicolumn{1}{c}{Model}&\multicolumn{1}{c}{CTG}&\multicolumn{1}{c}{Qual}&\multicolumn{1}{c}{ CTG}&\multicolumn{1}{c}{Qual}&\multicolumn{1}{c}{CTG}&\multicolumn{1}{c}{Qual}&\multicolumn{1}{c}{CTG}&\multicolumn{1}{c|}{Qual}&\multicolumn{1}{c}{CTG}&\multicolumn{1}{c}{Qual}\\ \hline
gpt-4-1106-preview & \multicolumn{1}{W@{}}{0.978}   & \multicolumn{1}{@{}W}{1.000}  & 0.505 & \multicolumn{1}{@{}W}{1.000} & \multicolumn{1}{W@{}}{0.969} & 0.995  & \multicolumn{1}{W@{}}{0.884} & \multicolumn{1}{@{}W|}{1.000} & \multicolumn{1}{W@{}}{0.834}   & \multicolumn{1}{@{}W}{0.999} \\
gpt-3.5-turbo-0125 & 0.335 & 0.989   & \multicolumn{1}{W@{}}{0.594} & 0.996   & 0.902 & \multicolumn{1}{@{}W}{0.996}  & 0.825 & 0.984 & 0.664  & 0.991  \\
gemini-pro  & 0.895   & 0.967  & 0.213 & 0.976 & 0.809  & 0.953   & 0.778  & 0.958  & 0.674  & 0.963 \\\hline
ca/calm2-7b-chat      & 0.293   & 0.909 & 0.176 & 0.973 & 0.373  & 0.956 & 0.673  & 0.922  & 0.379   & 0.940 \\
elyza/llama2-7b-instr  & 0.107   & 0.940 & 0.298 & 0.958 & 0.587  & 0.916 & 0.714  & 0.947 & 0.426   & 0.940 \\
line/llm-3.6b-instr  & 0.809   & 0.867   & 0.240 & 0.809 & 0.200  & 0.733 & 0.822  & 0.784 & 0.518   & 0.798   \\
matsuo/wl-10b-instr & 0.900   & 0.600 & 0.313 & 0.807 & 0.331  & 0.569 & 0.744  & 0.575  & 0.572   & 0.638  \\
rinna/youri-7b-chat & 0.322   & 0.593  & 0.187 & 0.795 & 0.471  & 0.640 & 0.836  & 0.662 & 0.454   & 0.673  \\
stabilityai/lm-instr-7b      & 0.296   & 0.795 & 0.287 & 0.773 & 0.500  & 0.729 & 0.715  & 0.778 & 0.449   & 0.769 \\
\hline
\end{tabular*}
\caption{Evaluation results of controllability (CTG) and quality (Qual) of generation in \textbf{pros \& cons generation.}}
\vspace{-3mm}
\label{tab:merideme-quality-result}
\end{center}
\end{table*}





Tables \ref{tab:summary-quality-result}, \ref{tab:ad-generation-quality-result}, and \ref{tab:merideme-quality-result} present the controllability performance and generated text quality for each model across the three tasks.

Overall, multilingual models GPT-3.5, GPT-4, and Gemini-Pro have demonstrated high performance in controllability and generation quality.
Notably, GPT-4 achieved the highest average scores across almost all tasks in each category.
In contrast, while the Japanese models exhibit high controllability in certain aspects, they generally display a significant performance gap compared to multilingual models like GPT-4.
In particular, in ca/calm2-7b-chat and elyza/llama2-7b-instr, the controllability scores are all low, despite the relatively high quality scores of approximately 0.7 - 0.9 for some tasks.
This result indicates that the evaluation of this benchmark provides insights that cannot be obtained by evaluating the content of the generated texts alone.

Also, all models scored low in regards to the \textsc{C-count} constraints. 
This finding suggests that language model tokenizers, which operate not on a character level but on a token level, may struggle with understanding constraints related to \textsc{C-count}.
Focusing on the \textsc{Keyword} and \textsc{P-word} within Japanese models, we observed that the trend in superiority and inferiority of the controllability scores for \textsc{Keyword} and \textsc{P-word} aligns across all Japanese models. 
This finding suggests that Japanese models may not comprehensively understand the differences between affirmative and negative expressions.
When comparing the same controllability aspects across the three tasks, significant score differences appear in the \textsc{Format}, \textsc{Keyword}, and \textsc{P-word} across all Japanese models. 
These results suggest that this benchmark can indicate a variance in difficulty levels for the exact controllability, depending on the task.

Moreover, despite rinna/youri-7b-chat achieving a high controllability score of 0.800 in \textsc{Format} constraints for the ad text generation, it demonstrates a considerably lower performance with a score of 0.320 in the quality of generation. 
This result implies that while rinna/youri-7b-chat exhibits adequate \textsc{Format} proficiency in generating ad text, its capacity to produce compelling ad text is somewhat lacking.
For instance, in response to the prompt in Figure \ref{fig:ad_generation_sample}, rinna/youri-7b-chat produced ``車の保険 (car insurance)'' as an output. Despite meeting the required conditions, it was considered inappropriate as an ad text title.
This result indicates the importance of simultaneously evaluating the controllability of LLMs and quality of the generated texts.
\section{Discussion}
\paragraph{Validity of quality evaluation by GPT-4} 
We manually conducted a similar quality evaluation for each task to confirm the validity of the quality evaluation of the generated texts from LLMs using GPT-4.
\begin{table*}[!t]
\begin{center}
\begin{tabular}{lrr}
\hline
\multicolumn{1}{c}{Task} & \multicolumn{1}{c}{\begin{tabular}[c]{@{}c@{}}GPT-4 vs Human \\ (Cohen's Kappa)\end{tabular}} & \multicolumn{1}{c}{\begin{tabular}[c]{@{}c@{}}Human Agreement\\ (Fleiss'Kappa)\end{tabular}} \\ \hline
Summarization   & 0.410  & 0.259 \\
Ad Text Generation      & 0.308  & 0.277 \\
Pros \& Cons Generation & 0.716  & 0.613 \\ \hline
\end{tabular}
\caption{Comparison of agreement levels in 
the evaluation of quality of generation.}
\label{tab:discussion:kappa-result}
\end{center}
\end{table*}

\begin{table*}[!t]
\begin{center}
\begin{tabular}{ccc}
\hline
Task & Exact Match (acc)   & Levenshtein based similarity \\\hline
\multicolumn{1}{l}{Summarization} & \multicolumn{1}{r}{0.863} & \multicolumn{1}{r}{0.960} \\ 
\multicolumn{1}{l}{Ad Text Generation} & \multicolumn{1}{r}{0.754} & \multicolumn{1}{r}{0.920}  \\
\multicolumn{1}{l}{Pros \& Cons Generation} & \multicolumn{1}{r}{0.827} & \multicolumn{1}{r}{0.948}\\
\hline
\end{tabular}
\caption{Agreement between sentences after manual removal and sentences after removal by GPT-4.}
\vspace{-3mm}
\label{tab:discussion:hf-remove}
\end{center}
\end{table*}
We used 48 samples from the summarization and 60 samples from the ad text generation and pros \&cons generation tasks corresponding to 10\% of the total data for each task.
We annotated the evaluation results against the generated texts by each model using the acquired samples as input.
Annotation was conducted by five annotation experts affiliated with our organization, and the majority vote results were considered human answers.
Furthermore, we checked the agreement of the answers among the five experts to confirm that the quality evaluation of the generated texts was a task of appropriate difficulty for people.

As shown in Table \ref{tab:discussion:kappa-result}, the agreement between GPT-4 and human evaluators, as well as the one between human evaluators, was substantial in the pros \& cons generation. However, it was moderate in the summarization and ad text generation.
This suggests that evaluating the quality of the summarization and ad text generation is more challenging than that for the pros \& cons generation.
Furthermore, even in the summarization, which is the most common among the three NLP tasks, various evaluation methods~\cite{song-etal-2024-finesure,shakil2024evaluatingtextsummariesgenerated} continued to be proposed.
This indicates that quality evaluation remains a highly complex issue. 
Therefore, further development of the quality evaluation methods is particularly needed for ad text generation and summarization.
We consider the evaluation by GPT-4 as a preliminary step toward robust quality evaluation.

\paragraph{Validity of the removal of unnecessary sentences using GPT-4} 
We manually removed the text using crowdsourcing and compared the results of the two methods to confirm that GPT-4 properly removed unnecessary explanatory text.
We collect the same number of data as in the ``Validity of quality evaluation by GPT-4'' for each task and use it for manual assessment.
For each sample, one person is tasked with extracting the text that should be removed if it is at the beginning or the end of a sentence.
The results of manual removal are compared with those of removal by GPT-4 from two aspects: the percentage of cases where the beginning and the last ten characters of two sentences match exactly (Exact Match) and the similarity based on the Levenshtein distance.

As shown in Table \ref{tab:discussion:hf-remove}. 
The overall similarity is very high, indicating that the \textsc{C-count}, \textsc{Keyword}, and \textsc{P-word} do not negatively impact the evaluation. 
The comparison by matching the beginning and the end of sentences also yields high scores. 
Even in the case of ad text generation, where accuracy is relatively low, high scores were obtained for the \textsc{Format} in GPT-4 and 3.5 as indicated in Table \ref{tab:ad-generation-quality-result}. 
Hence, we can assert that there are no negative impacts, such as instances where responses that meet the \textsc{Format} are inaccurately addressed.
We can conclude that GPT-4 is sufficient to removes unnecessary explanatory text.

\section{Conclusion}
We constructed the LCTG Bench, a benchmark for evaluating the controllability of LLMs, which consists of three types of tasks and four perspectives of controllability.
This dataset is the first Japanese LLM benchmark that allows for robust evaluation of the controllability across various tasks.
We evaluated the controllability of nine Japanese LLMs and three high-performance multilingual models, such as GPT-4.
Overall, GPT-4 demonstrated superior performance both in terms of controllability and quality, revealing a significant performance gap between GPT-4 and Japanese LLMs.
We also conducted a detailed analysis of each controllability perspective.
Notably, the performance of character count control (\textsc{C-COUNT}) was low, even for GPT-4, suggesting that future improvements are needed in this perspective.
\footnote{The LCTG Bench is available at this page:  \url{https://github.com/CyberAgentAILab/LCTG-Bench}.}
\newpage

\bibliography{reference}

\begin{thebibliography}{30}
\providecommand{\natexlab}[1]{#1}

\bibitem[{Black et~al.(2022)Black, Biderman, Hallahan, Anthony, Gao, Golding, He, Leahy, McDonell, Phang, Pieler, Prashanth, Purohit, Reynolds, Tow, Wang, and Weinbach}]{black2022gptneox20b}
Sid Black, Stella Biderman, Eric Hallahan, Quentin Anthony, Leo Gao, Laurence Golding, Horace He, Connor Leahy, Kyle McDonell, Jason Phang, Michael Pieler, USVSN~Sai Prashanth, Shivanshu Purohit, Laria Reynolds, Jonathan Tow, Ben Wang, and Samuel Weinbach. 2022.
\newblock \href {https://arxiv.org/abs/2204.06745} {{GPT}-{N}eo{X}-{20B}: An open-source autoregressive language model}.
\newblock Abs/2204.06745.

\bibitem[{Chan et~al.(2023)Chan, Chen, Su, Yu, Xue, Zhang, Fu, and Liu}]{chan2023chateval}
Chi-Min Chan, Weize Chen, Yusheng Su, Jianxuan Yu, Wei Xue, Shanghang Zhang, Jie Fu, and Zhiyuan Liu. 2023.
\newblock \href {https://arxiv.org/abs/2308.07201} {Chat{E}val: Towards better {LLM}-based evaluators through multi-agent debate}.
\newblock \emph{Preprint}, arXiv:2308.07201.

\bibitem[{Fu et~al.(2023)Fu, Ng, Jiang, and Liu}]{fu2023gptscore}
Jinlan Fu, See-Kiong Ng, Zhengbao Jiang, and Pengfei Liu. 2023.
\newblock {GPTS}core: Evaluate as you desire.
\newblock \emph{arXiv:2302.04166}.

\bibitem[{Gao et~al.(2021)Gao, Tow, Biderman, Black, DiPofi, Foster, Golding, Hsu, McDonell, Muennighoff, Phang, Reynolds, Tang, Thite, Wang, Wang, and Zou}]{eval-harness}
Leo Gao, Jonathan Tow, Stella Biderman, Sid Black, Anthony DiPofi, Charles Foster, Laurence Golding, Jeffrey Hsu, Kyle McDonell, Niklas Muennighoff, Jason Phang, Laria Reynolds, Eric Tang, Anish Thite, Ben Wang, Kevin Wang, and Andy Zou. 2021.
\newblock \href {https://doi.org/10.5281/zenodo.5371628} {A framework for few-shot language model evaluation}.

\bibitem[{Gemini-Team et~al.(2024)Gemini-Team, Anil, Borgeaud, Alayrac, Yu, Soricut, Schalkwyk, Dai, and et~al.}]{geminiteam2024geminifamilyhighlycapable}
Gemini-Team, Rohan Anil, Sebastian Borgeaud, Jean-Baptiste Alayrac, Jiahui Yu, Radu Soricut, Johan Schalkwyk, Andrew~M Dai, and Anja~Hauth et~al. 2024.
\newblock \href {https://arxiv.org/abs/2312.11805} {Gemini: A family of highly capable multimodal models}.
\newblock Abs/2312.11805.

\bibitem[{Hasan et~al.(2021)Hasan, Bhattacharjee, Islam, Mubasshir, Li, Kang, Rahman, and Shahriyar}]{hasan-etal-2021-xl}
Tahmid Hasan, Abhik Bhattacharjee, Md.~Saiful Islam, Kazi Mubasshir, Yuan-Fang Li, Yong-Bin Kang, M.~Sohel Rahman, and Rifat Shahriyar. 2021.
\newblock \href {https://doi.org/10.18653/v1/2021.findings-acl.413} {{XL}-{S}um: Large-scale multilingual abstractive summarization for 44 languages}.
\newblock In \emph{Findings of the Association for Computational Linguistics: ACL-IJCNLP 2021}, pages 4693--4703, Online. Association for Computational Linguistics.

\bibitem[{He et~al.(2024)He, Zeng, Huang, Chen, Xiao, He, Zhou, Chen, Wang, Huang, Ye, Li, Chen, Zhang, Gu, Liang, and Xiao}]{he2024celloAAAI}
Qianyu He, Jie Zeng, Wenhao Huang, Lina Chen, Jin Xiao, Qianxi He, Xunzhe Zhou, Lida Chen, Xintao Wang, Yuncheng Huang, Haoning Ye, Zihan Li, Shisong Chen, Yikai Zhang, Zhouhong Gu, Jiaqing Liang, and Yanghua Xiao. 2024.
\newblock Can large language models understand real-world complex instructions?
\newblock In \emph{Proceedings of the 38th AAAI Conference on Artificial Intelligence}, Vancouver, Canada.

\bibitem[{Hendrycks et~al.(2021)Hendrycks, Burns, Basart, Zou, Mazeika, Song, and Steinhardt}]{hendrycks2021measuring}
Dan Hendrycks, Collin Burns, Steven Basart, Andy Zou, Mantas Mazeika, Dawn Song, and Jacob Steinhardt. 2021.
\newblock \href {https://arxiv.org/abs/2009.03300} {Measuring massive multitask language understanding}.
\newblock Abs/2009.03300.

\bibitem[{Jiang et~al.(2023)Jiang, Wang, Zeng, Zhong, Li, Mi, Shang, Jiang, Liu, and Wang}]{jiang2023followbench}
Yuxin Jiang, Yufei Wang, Xingshan Zeng, Wanjun Zhong, Liangyou Li, Fei Mi, Lifeng Shang, Xin Jiang, Qun Liu, and Wei Wang. 2023.
\newblock \href {https://arxiv.org/abs/2310.20410} {{F}ollow{B}ench: A multi-level fine-grained constraints following benchmark for large language models}.
\newblock Abs/2310.20410.

\bibitem[{Jing et~al.(2023)Jing, Jin, Hu, Qiu, Wang, Wang, and Xiong}]{jing2023followeval}
Yimin Jing, Renren Jin, Jiahao Hu, Huishi Qiu, Xiaohua Wang, Peng Wang, and Deyi Xiong. 2023.
\newblock \href {https://arxiv.org/abs/2311.09829} {{F}ollow{E}val: A multi-dimensional benchmark for assessing the instruction-following capability of large language models}.
\newblock Abs/2311.09829.

\bibitem[{Kurihara et~al.(2022)Kurihara, Kawahara, and Shibata}]{kurihara-etal-2022-jglue}
Kentaro Kurihara, Daisuke Kawahara, and Tomohide Shibata. 2022.
\newblock \href {https://aclanthology.org/2022.lrec-1.317} {{JGLUE}: {J}apanese general language understanding evaluation}.
\newblock In \emph{Proceedings of the Thirteenth Language Resources and Evaluation Conference}, pages 2957--2966, Marseille, France. European Language Resources Association.

\bibitem[{Liu et~al.(2023)Liu, Fabbri, Chen, Zhao, Han, Joty, Liu, Radev, Wu, and Cohan}]{liu2023benchmarking}
Yixin Liu, Alexander~R. Fabbri, Jiawen Chen, Yilun Zhao, Simeng Han, Shafiq Joty, Pengfei Liu, Dragomir Radev, Chien-Sheng Wu, and Arman Cohan. 2023.
\newblock \href {https://arxiv.org/abs/2311.09184} {Benchmarking generation and evaluation capabilities of large language models for instruction controllable summarization}.
\newblock Abs/2311.09184.

\bibitem[{Maynez et~al.(2023)Maynez, Agrawal, and Gehrmann}]{maynez-etal-2023-benchmarking}
Joshua Maynez, Priyanka Agrawal, and Sebastian Gehrmann. 2023.
\newblock \href {https://doi.org/10.18653/v1/2023.acl-long.511} {Benchmarking large language model capabilities for conditional generation}.
\newblock In \emph{Proceedings of the 61st Annual Meeting of the Association for Computational Linguistics (Volume 1: Long Papers)}, pages 9194--9213, Toronto, Canada. Association for Computational Linguistics.

\bibitem[{Mita et~al.(2023)Mita, Murakami, Kato, and Zhang}]{mita2023}
Masato Mita, Soichiro Murakami, Akihiko Kato, and Peinan Zhang. 2023.
\newblock \href {https://arxiv.org/abs/2309.12030} {{CAMERA}: A multimodal dataset and benchmark for ad text generation}.
\newblock Abs/2309.12030.

\bibitem[{Mizrahi et~al.(2023)Mizrahi, Kaplan, Malkin, Dror, Shahaf, and Stanovsky}]{mizrahi2023state}
Moran Mizrahi, Guy Kaplan, Dan Malkin, Rotem Dror, Dafna Shahaf, and Gabriel Stanovsky. 2023.
\newblock \href {https://arxiv.org/abs/2401.00595} {State of what art? {A} call for multi-prompt {LLM} evaluation}.
\newblock Abs/2401.00595.

\bibitem[{OpenAI(2023)}]{OpenAI_GPT4_2023}
OpenAI. 2023.
\newblock \href {https://arxiv.org/abs/2303.08774} {{GPT}-4 technical report}.
\newblock \emph{ArXiv}, abs/2303.08774.

\bibitem[{Ouyang et~al.(2023)Ouyang, Zhang, Harman, and Wang}]{ouyang2023llm}
Shuyin Ouyang, Jie~M. Zhang, Mark Harman, and Meng Wang. 2023.
\newblock \href {https://arxiv.org/abs/2308.02828} {{LLM} is like a box of chocolates: the non-determinism of chatgpt in code generation}.
\newblock Abs/2308.02828.

\bibitem[{Shakil et~al.(2024)Shakil, Mahi, Nguyen, Ortiz, and Mardini}]{shakil2024evaluatingtextsummariesgenerated}
Hassan Shakil, Atqiya~Munawara Mahi, Phuoc Nguyen, Zeydy Ortiz, and Mamoun~T. Mardini. 2024.
\newblock \href {https://arxiv.org/abs/2405.04053} {Evaluating text summaries generated by large language models using {O}pen{AI}'s {GPT}}.
\newblock Abs/2405.04053.

\bibitem[{Shi et~al.(2022)Shi, Suzgun, Freitag, Wang, Srivats, Vosoughi, Chung, Tay, Ruder, Zhou, Das, and Wei}]{shi2022language}
Freda Shi, Mirac Suzgun, Markus Freitag, Xuezhi Wang, Suraj Srivats, Soroush Vosoughi, Hyung~Won Chung, Yi~Tay, Sebastian Ruder, Denny Zhou, Dipanjan Das, and Jason Wei. 2022.
\newblock \href {https://arxiv.org/abs/2210.03057} {Language models are multilingual chain-of-thought reasoners}.
\newblock Abs/2210.03057.

\bibitem[{Someya et~al.(2023)Someya, Sugimoto, and Oseki}]{someya2023jcola}
Taiga Someya, Yushi Sugimoto, and Yohei Oseki. 2023.
\newblock \href {https://arxiv.org/abs/2309.12676} {{JC}o{LA}: Japanese corpus of linguistic acceptability}.

\bibitem[{Song et~al.(2024)Song, Su, Shalyminov, Cai, and Mansour}]{song-etal-2024-finesure}
Hwanjun Song, Hang Su, Igor Shalyminov, Jason Cai, and Saab Mansour. 2024.
\newblock \href {https://aclanthology.org/2024.acl-long.51} {{F}ine{S}ur{E}: Fine-grained summarization evaluation using {LLM}s}.
\newblock In \emph{Proceedings of the 62nd Annual Meeting of the Association for Computational Linguistics (Volume 1: Long Papers)}, pages 906--922, Bangkok, Thailand. Association for Computational Linguistics.

\bibitem[{Sun et~al.(2023{\natexlab{a}})Sun, Tian, Zhou, Xu, Hu, Gupta, Wieting, Peng, and Ma}]{sun2023evaluating}
Jiao Sun, Yufei Tian, Wangchunshu Zhou, Nan Xu, Qian Hu, Rahul Gupta, John~Frederick Wieting, Nanyun Peng, and Xuezhe Ma. 2023{\natexlab{a}}.
\newblock \href {https://arxiv.org/abs/2310.14542} {Evaluating large language models on controlled generation tasks}.
\newblock Abs/2310.14542.

\bibitem[{Sun et~al.(2023{\natexlab{b}})Sun, Yan, Ma, Ren, Yin, and Ren}]{sun2023chatgpt}
Weiwei Sun, Lingyong Yan, Xinyu Ma, Pengjie Ren, Dawei Yin, and Zhaochun Ren. 2023{\natexlab{b}}.
\newblock Is chatgpt good at search? investigating large language models as re-ranking agent.
\newblock \emph{arXiv preprint arXiv:2304.09542}.

\bibitem[{Suzgun et~al.(2023)Suzgun, Scales, Sch{\"a}rli, Gehrmann, Tay, Chung, Chowdhery, Le, Chi, Zhou, and Wei}]{suzgun-etal-2023-challenging}
Mirac Suzgun, Nathan Scales, Nathanael Sch{\"a}rli, Sebastian Gehrmann, Yi~Tay, Hyung~Won Chung, Aakanksha Chowdhery, Quoc Le, Ed~Chi, Denny Zhou, and Jason Wei. 2023.
\newblock \href {https://doi.org/10.18653/v1/2023.findings-acl.824} {Challenging {BIG}-bench tasks and whether chain-of-thought can solve them}.
\newblock In \emph{Findings of the Association for Computational Linguistics: ACL 2023}, pages 13003--13051, Toronto, Canada. Association for Computational Linguistics.

\bibitem[{Suzuki et~al.(2020)Suzuki, Matsuda, Okazaki, and Inui}]{Kurihara_nlp2020}
Masatoshi Suzuki, Koji Matsuda, Naoaki Okazaki, and Kentaro Inui. 2020.
\newblock {JAQKET}: Quiz wo daizai ni shita nihongo qa dataset no kochiku.
\newblock In \emph{NLP2020}.
\newblock In Japanese.

\bibitem[{Tikhonov and Ryabinin(2021)}]{tikhonov-ryabinin-2021-heads}
Alexey Tikhonov and Max Ryabinin. 2021.
\newblock \href {https://doi.org/10.18653/v1/2021.findings-acl.310} {{I}t{'}s {A}ll in the {H}eads: {U}sing {A}ttention {H}eads as a {B}aseline for {C}ross-{L}ingual {T}ransfer in {C}ommonsense {R}easoning}.
\newblock In \emph{Findings of the Association for Computational Linguistics: ACL-IJCNLP 2021}, pages 3534--3546, Online. Association for Computational Linguistics.

\bibitem[{Touvron et~al.(2023)Touvron, Lavril, Izacard, Martinet, Lachaux, Lacroix, Rozière, Goyal, Hambro, Azhar, Rodriguez, Joulin, Grave, and Lample}]{touvron2023llama}
Hugo Touvron, Thibaut Lavril, Gautier Izacard, Xavier Martinet, Marie-Anne Lachaux, Timothée Lacroix, Baptiste Rozière, Naman Goyal, Eric Hambro, Faisal Azhar, Aurelien Rodriguez, Armand Joulin, Edouard Grave, and Guillaume Lample. 2023.
\newblock \href {https://arxiv.org/abs/2302.13971} {{LL}a{MA}: Open and efficient foundation language models}.
\newblock Abs/2302.13971.

\bibitem[{Yao et~al.(2023)Yao, Chen, Hanjie, Yang, and Narasimhan}]{yao2023collie}
Shunyu Yao, Howard Chen, Austin~W. Hanjie, Runzhe Yang, and Karthik Narasimhan. 2023.
\newblock \href {https://arxiv.org/abs/2307.08689} {{COLLIE}: Systematic construction of constrained text generation tasks}.
\newblock Abs/2307.08689.

\bibitem[{Zheng et~al.(2023)Zheng, Chiang, Sheng, Zhuang, Wu, Zhuang, Lin, Li, Li, Xing, Zhang, Gonzalez, and Stoica}]{zheng2023judging}
Lianmin Zheng, Wei-Lin Chiang, Ying Sheng, Siyuan Zhuang, Zhanghao Wu, Yonghao Zhuang, Zi~Lin, Zhuohan Li, Dacheng Li, Eric.~P Xing, Hao Zhang, Joseph~E. Gonzalez, and Ion Stoica. 2023.
\newblock \href {https://arxiv.org/abs/2306.05685} {Judging {LLM}-as-a-{J}udge with mt-bench and chatbot arena}.
\newblock Abs/2306.05685.

\bibitem[{Zhou et~al.(2023)Zhou, Lu, Mishra, Brahma, Basu, Luan, Zhou, and Hou}]{zhou2023instructionfollowing}
Jeffrey Zhou, Tianjian Lu, Swaroop Mishra, Siddhartha Brahma, Sujoy Basu, Yi~Luan, Denny Zhou, and Le~Hou. 2023.
\newblock \href {https://arxiv.org/abs/2311.07911} {Instruction-following evaluation for large language models}.
\newblock Abs/2311.07911.

\end{thebibliography}

\appendix
\onecolumn
\section{Collecting Templates of Conditional Sentences through Crowdsourcing}
\label{appendix:cs_example}
As discussed in Section \ref{ssec:benchmark:eval_item}, we collected templates for conditional statements for each of the perspectives of controllability: \textsc{Format}, \textsc{C-count}, \textsc{Keyword}, \textsc{P-word}.
The templates for the conditional statements were collected using crowdsourcing, which involved rewriting a given conditional statement to retain its original meaning. Finally, we manually removed any inappropriate examples, such as templates with different meanings. 
The number of templates eventually adopted is shown in Table \ref{tab:templates_num}, and the examples of collected templates are presented below.

\textbf{\textsc{C-count}}:  「XXX-YYY文字で要約して」, 「XXX-YYY文字でまとめること」, 「XXX文字以上、XXX文字以下で要約」 ... (All these instructions are to perform the task within the range of XXX-YYY characters.)

\textbf{\textsc{Keyword}}:  「XXX」という単語を含める, XXXという言葉を使ってください。... (All these instructions are to perform the task in a way that the generated result includes the word `XXX.')

\textbf{\textsc{P-word}}: 「XXX」という単語は入れない, 「XXX」という言葉は使用不可 ... (All these instructions are to perform the task in a way that the generated result does not include the word `XXX.')

\section{Conditional statement of `Format'}
\label{appendix:format_prompt_ex}
In the conditional statements of `Format', we used predetermined sentences. The conditional statements of `Format' used in each of the three tasks are presented below.

\textbf{Summarization}: 文章の要約結果のみを出力し、要約結果の前後に説明文などは付与しないでください。(Please output only the summary results and refrain from adding any explanatory sentences before or after the summary.
)

\textbf{Ad Text Generation}:  広告文のタイトルのみを出力し、広告文のタイトルの前後に説明文などは付与しないでください。
(Please ensure to output only the title of the advertisement, without adding any explanatory text before or after the title.)

\textbf{Pros \& Cons Generation}:  メリットデメリットに関する回答の前後に「〇〇するメリットとデメリットは以下です。」「以上が〇〇するメリットとデメリットです。」などの説明文を付与しないでください。
(Regarding the pros and cons, please avoid adding explanatory sentences such as ``The advantages and disadvantages of doing 〇〇 are as follows.'' and ``The above are the advantages and disadvantages of doing 〇〇.'' before and after the answer. )

\section{Models and configuration of hyperparameters}
\label{appendix:hypara}
The models used in the experiment and the settings for various hyperparameters are shown in Table \ref{tbl:llm_list_ap}.

\section{Generated result with unnecessary text}
\label{appendix:with_hf_example}
An example of unnecessary explanatory text included at the beginning and end of a sentence in the generated result is shown in Figure \ref{fig:with_hf_generation_result}.

\section{Prompt for removing unnecessary explanatory phrases}
\label{appendix:hf_remover_prompt}
The prompt inputted into GPT-4 for removing unnecessary explanatory phrases included in the generated texts of LLM is shown in Figures \ref{fig:summary_hf_remove_prompt}, \ref{fig:ad_generation_hf_remove_prompt} and \ref{fig:merideme_generation_hf_remove_prompt}.

\section{Prompt used for evaluating the quality of generation}
\label{appendix:quality_check_prompt}
The prompts inputted into GPT-4 for evaluating the quality of LLM generation are shown in Figures \ref{fig:summary_quality_check_prompt}, \ref{fig:ad_generation_quality_check_prompt} and \ref{fig:merideme_generation_quality_check_prompt}.

\onecolumn
\begin{table*}[!h]
\centering
\begin{tabular}{lrrr}
\hline
& \multicolumn{1}{c}{\textsc{C-count}} & \multicolumn{1}{c}{\textsc{Keyword}} & \multicolumn{1}{c}{\textsc{P-word}} \\
\hline
Summarization& 107  & 93  & 96 \\
Ad Text Generation      & 105  & 92  & 95 \\
Pros \& Cons Generation & 132  & 150 & 147 \\
\hline
\end{tabular}
\caption{The number of templates in each task and perspectives of controllability.}
\label{tab:templates_num}
\end{table*}

\begin{table}[!h]
\small
\begin{center}
\begin{tabular}{@{}l@{}crrr}
\hline
\multicolumn{1}{c}{Model} & Base Model & \multicolumn{1}{c}{max\_new\_tokens} & \multicolumn{1}{c}{temperature} & \multicolumn{1}{c}{top\_p} \\ \hline
gpt-4-1106-preview & - & - & 1.0 & 1.0 \\
gpt-3.5-turbo-0125 & - & - & 1.0 & 1.0 \\
gemini-pro & - & 8,092 & 0.9 & 1.0 \\
ca/calm2-7b-chat\tablefootnote{\url{https://huggingface.co/cyberagent/calm2-7b-chat}} & Llama 2 & 4,096 & 0.9 & - \\
elyza/llama2-7b-instr\tablefootnote{\url{https://huggingface.co/elyza/ELYZA-japanese-Llama-2-7b-fast-instruct}} & Llama 2 & 4,096 & 0.9 & - \\
line/llm-3.6b-instr\tablefootnote{\url{https://huggingface.co/line-corporation/japanese-large-lm-3.6b-instruction-sft}} & GPT-NeoX & 4,096 & 0.8 & 0.9 \\
matsuo/wl-10b-instr\tablefootnote{\url{https://huggingface.co/matsuo-lab/weblab-10b-instruction-sft}} & GPT-NeoX & 4,096 & 0.8 & 0.95 \\
rinna/youri-7b-chat\tablefootnote{\url{https://huggingface.co/rinna/youri-7b-chat}} & Llama 2 & 4,096 & - & - \\
stabilityai/lm-instr-7b\tablefootnote{\url{https://huggingface.co/stabilityai/japanese-stablelm-instruct-gamma-7b}} & Mistral & 4,096 & 0.5 & 0.95 \\
\hline
\end{tabular}
\caption{The list of LLMs used in our experiments and the configuration of hyperparameters.}
\label{tbl:llm_list_ap}
\end{center}
\end{table}

\begin{figure}[h]
\small
\begin{center}
\begin{tabular}{p{13.1cm}}\\\hline
承知しました。与えられた文章に企業という言葉を使わないタイトルを作成します。\\
タイトル: 40の条件であなただけの営業リストを作成\\
このタイトルは、与えられた文章に企業という言葉を使わず、営業リストを作成するサービスをアピールする内容となっています。\\
\footnotesize (\textit{Sure. I will create a title for the given text \footnotesize without using the word ``company''.} \\
\footnotesize \textit{Title: Creating Your Unique Sales List with 40 Criteria} \\
\footnotesize \textit{This title does not use the word "company" from the given text, and emphasizes a service that creates a sales list.})
\\\hline
\end{tabular}
\caption{An Example of LLM output in ad text generation that includes irrelevant explanations at the beginning and end of sentences: the explanations are included in the character count, making it impossible to measure the appropriate number of characters for task response.}
\label{fig:with_hf_generation_result}
\end{center}
\end{figure}

\begin{figure}[t]
\small
\begin{center}

\begin{tabular}{p{13.1cm}}\\\hline
以下に提示している文章は、ある文章を生成AIを用いて要約した出力結果です。\\
出力には「要約」あるいはそれに類する単語を含むような文として、「以下の文章を要約します。」「【要約】」などの冒頭の説明文や「以上が要約結果になります。」などの文末の説明文が入っていることがあります。また、英語でこれらの説明文が与えられることもあります。\\
提示した文章に上記で述べた説明文が含まれていない場合には提示した文章をそのまま出力し、上記で述べた説明文が含まれている場合は提示した文章から説明文を除去したものを抜き出してください。文章の中間部分を編集する必要は一切ありません。文が入っていることがあります。また、英語でこれらの説明文が与えられることもあります。\\
\footnotesize (\textit{The text provided below is an output generated by a summarization AI.} \\
\footnotesize \textit{In the output, there may be sentences that include words such as 'summary' or similar, serving as introductory or concluding explanations, like "The following text will be summarized." or "This concludes the summary.", among others. Additionally, these explanatory sentences might also be given in English. }\\
\footnotesize \textit{If the provided text does not contain the aforementioned explanatory sentences, please output the text as is. If it does contain these explanatory sentences, extract the text by removing these explanations. There is no need to edit the middle part of the text. Sentences may be included. These descriptions may also be given in English.})\\

[文章]\\
\footnotesize ([\textit{Text}])\\
\{生成結果\}\\
(\{\textit{generated\_result}\})
\\\hline
\end{tabular}
\caption{Prompt used for removing unnecessary explanatory phrases in summarization.}
\label{fig:summary_hf_remove_prompt}
\end{center}
\end{figure}

\begin{figure}[h]
\small
\begin{center}
\begin{tabular}{p{13.1cm}}\\\hline
以下に提示している文章は、ある文章を元に作成した広告文のタイトルです。\\
出力には「広告文：」や「広告文を作成します」などの冒頭の接頭辞や説明文、「作成しました。」「このタイトルは、、」などの接尾辞やタイトルの後ろの説明文が含まれていることがあります。\\
提示した文章に上記で述べた説明文や接頭辞、接尾辞が含まれていない場合には、提示した文章をそのまま出力してください。「」や**などの記号で囲われている事例の場合、記号は全て残したまま出力してください。\\
上記で述べた説明文が含まれている場合は提示した文章から説明文や接頭辞、接尾辞を除去したものを抜き出してください。冒頭、末尾以外の中間部分を編集する必要は一切ありません。新しく文字を追加などをしないでください。\\
\footnotesize (\textit{The following text is the title of an advertisement created based on a certain article.} \\
\footnotesize \textit{In the output, there may be initial prefixes or explanatory text such as "Advertisement:", "Creating an advertisement", and suffixes or explanatory text following the title such as "Created.", "This title is,,".} \\
\footnotesize \textit{If the text provided does not include the explanatory text or prefixes and suffixes mentioned above, please output the provided text as is. For examples enclosed in symbols such as "" or **, please leave all symbols intact in the output.} \\
\footnotesize \textit{If the explanatory text mentioned above is included, please extract the text from the provided text by removing the explanatory text or prefixes and suffixes. There is absolutely no need to edit the middle part other than the beginning and end. Please do not add new characters.})\\
\\

[文章]\\
\footnotesize ([\textit{Text}])\\
\{生成結果\}\\
(\{\textit{generated\_result}\})
\\\hline
\end{tabular}
\caption{Prompt used for removing unnecessary explanatory phrases in ad text generation.}
\label{fig:ad_generation_hf_remove_prompt}
\end{center}
\end{figure}

\begin{figure}[h]
\small
\begin{center}
\begin{tabular}{p{13.1cm}}\\\hline
以下に提示している文章は、ある事象・事物についてのメリットとデメリットを生成AIに回答してもらった出力結果です。\\
文章の冒頭や末尾に「そこで、メリットとデメリットをご紹介いたします。」「あなたのご質問にお答えいたします。」「以上が〇〇に関するメリット・デメリットです。」など内容と関係のない説明文が付与されている場合は、その説明文を除去して出力してください。ただし、文の一部は変更せずに、該当の文全体を除去してください。\\
上記のような説明文が付与されていない場合は、提示している文章をそのまま出力してください。\\
\footnotesize (\textit{The following text is the output result of an AI answering the merits and demerits of a certain event or thing.} \\
\footnotesize \textit{If there is unrelated explanatory text at the beginning or end of the sentence such as "Therefore, let me introduce the merits and demerits.", "I will answer your question.", "The above are the merits and demerits of 〇〇.", please remove that explanatory text and output. However, without changing any part of the sentence, please remove the entire relevant sentence.} \\
\footnotesize \textit{If no such explanatory text is attached, please output the text as it is presented.})\\
\\

[文章]\\
\footnotesize ([\textit{Text}])\\
\{生成結果\}\\
(\{\textit{generated\_result}\})
\\\hline
\end{tabular}
\caption{Prompt used for removing unnecessary explanatory phrases in pros \& cons Generation.}
\label{fig:merideme_generation_hf_remove_prompt}
\end{center}
\end{figure}

\begin{figure}[h]
\small
\begin{center}
\begin{tabular}{p{13.1cm}}\\\hline
以下に要約した文章とその要約元の文章が提示されています。\\
要約した文章は要約元の文章を適切に要約できているかを判断してください。\\
適切に要約できている場合は「適切」、適切に要約できていない場合は「不適切」と回答してください。\\
ただし、要約元の文章から断定できない情報が要約した文章に含まれている場合も「不適切」と回答してください。\\
「適切」「不適切」のいずれかのみを出力し、説明文などは付与しないでください。\\
\footnotesize (\textit{The following presents a summarized text and the original text from which it was summarized.} \\
\footnotesize \textit{Please judge whether the summarized text appropriately summarizes the original text.} \\
\footnotesize \textit{If it is appropriately summarized, answer "Appropriate." If it is not appropriately summarized, answer "Inappropriate."} \\
\footnotesize \textit{However, if the summarized text includes information that cannot be confirmed from the original text, also answer "Inappropriate."} \\
\footnotesize \textit{Please output only one of "Appropriate" or "Inappropriate," and do not add any explanatory text.})\\\\
【要約元の文章】\\
\footnotesize (【\textit{Original Text}】)\\
\{要約元の文章\}\\
\footnotesize \{\textit{original\_text}\}\\\\
【要約した文章】\\
\footnotesize (【\textit{Summarized Text}】)\\
\{生成結果\}\\
\footnotesize (\{\textit{generated\_result}\})
\\\hline
\end{tabular}
\caption{Prompt used for evaluating the quality of generation in summarization.}
\label{fig:summary_quality_check_prompt}
\end{center}
\end{figure}

\begin{figure}[h]
\small
\begin{center}
\begin{tabular}{p{13.1cm}}\\\hline
以下に、ランディングページの説明文とその説明文をもとに作成した1つの広告文のタイトルがあります。\\
説明文の内容に基づいているタイトルを作成できているかを判断してください。\\
適切に作成できている場合は「適切」、適切に作成できていない場合は「不適切」と回答してください。\\
ただし、説明文とタイトルが完全に一致している事例とタイトルとして長すぎる事例も「不適切」と回答してください。\\
「適切」「不適切」のいずれかのみを出力し、説明文などは付与しないでください。\\
\footnotesize \textit{Below is a description of the landing page and the title of one ad text created based on that description.} \\
\footnotesize \textit{Please judge whether the title is created based on the content of the explanatory text.} \\
\footnotesize \textit{If it is appropriately created, answer ”Appropriate”. If it is not appropriately created, answer ”Inappropriate.”} \\
\footnotesize \textit{However, please also answer "Inappropriate" for cases where the explanatory text and the title are completely identical and cases where the title is too long.} \\
\footnotesize \textit{Please output only one of "Appropriate" or "Inappropriate," and do not add any explanatory text.} )\\\\
【説明文】\\
\footnotesize (【\textit{Explanatory Text}】)\\
\{LPテキスト\}\\
\footnotesize (\{\textit{LP Text}\})\\\\
【広告文のタイトル】\\
\footnotesize (【\textit{Advertisement} Title】)\\
\{生成結果\}\\
\footnotesize (\{\textit{generated\_result}\})
\\\hline
\end{tabular}
\caption{Prompt used for evaluating the quality of generation in ad text generation.}
\label{fig:ad_generation_quality_check_prompt}
\end{center}
\end{figure}

\begin{figure}[h]
\small
\begin{center}
\begin{tabular}{p{13.1cm}}\\\hline
以下に提示している文章は、ある事象・事物についてのメリットとデメリットを生成AIに回答してもらった出力結果です。\\
出力結果が、メリット・デメリットの双方について言及できているか否かを回答してください。\\
言及できている場合は「適切」、言及できていない場合は「不適切」と回答してください。\\
「適切」「不適切」のいずれかのみを出力し、説明文などは付与しないでください。\\
\footnotesize (\textit{The following text is the output result of an AI answering the pros and cons of a certain event or thing.} \\
\footnotesize \textit{Please answer whether the output result can mention both pros and cons.} \\
\footnotesize \textit{If it can mention, answer "Appropriate". If it cannot mention, answer "Inappropriate".} \\
\footnotesize \textit{Please output only one of "Appropriate" or "Inappropriate", and do not add any explanatory text.})\\
\\
【文章】\\
\footnotesize (【\textit{Text}】)\\
\{生成結果\}\\
\footnotesize (\{\textit{generated\_result}\})
\\\hline
\end{tabular}
\caption{Prompt used for removing unnecessary explanatory phrases in pros \& cons generation.}
\label{fig:merideme_generation_quality_check_prompt}
\end{center}
\end{figure}

\end{document}